\documentclass[10pt]{article} 
\usepackage[preprint]{tmlr}

\usepackage{amsmath,amsfonts,bm}

\def\eqref#1{equation~\ref{#1}}

\def\1{\bm{1}}

\DeclareMathAlphabet{\mathsfit}{\encodingdefault}{\sfdefault}{m}{sl}
\SetMathAlphabet{\mathsfit}{bold}{\encodingdefault}{\sfdefault}{bx}{n}

\usepackage{hyperref}
\usepackage{url}
\usepackage{amssymb}
\usepackage{multirow}
\usepackage{graphicx}
\usepackage{comment}
\usepackage{dsfont}
\usepackage{wrapfig}
\title{Variational Augmented Invertible Koopman Autoencoder \\for probabilistic time series forecasting}

\author{\name Anthony Frion \email anthony.frion@hereon.de \\
      \addr Institute of coastal modelling\\
      Helmholtz-Zentrum Hereon
      \AND
      \name Lucas Drumetz \email lucas.drumetz@imt-atlantique.fr \\
      \addr IMT Atlantique \\
      Lab-STICC, UMR CNRS 6285, Brest, France
      \AND
      \name Guillaume Tochon \email guillaume.tochon@lrde.epita.fr \\
      \addr LRE EPITA, Le Kremlin-Bicêtre, France
      \AND
      \name Mauro Dalla Mura \email mauro.dalla-mura@gipsa-lab.grenoble-inp.fr\\
      \addr Université Grenobles Alpes \\
      Grenoble INP \\
      GIPSA-lab, Grenoble, France \\
      Institut Universitaire de France
      \AND
      \name Ali Can Bekar \email ali.bekar@hereon.de\\
      \addr Institute of coastal modelling\\
      Helmholtz-Zentrum Hereon
      \AND
      \name Abdeldjalil Aïssa El Bey \email abdeldjalil.aissaelbey@imt-atlantique.fr \\
      \addr IMT Atlantique \\
      Lab-STICC, UMR CNRS 6285, Brest, France
      }

\def\month{MM}  
\def\year{YYYY} 
\def\openreview{\url{https://openreview.net/forum?id=XXXX}} 

\begin{document}

\maketitle

\begin{abstract}
Neural Koopman autoencoder models have been shown to successfully build a latent embedding with linear dynamics for arbitrary dynamical systems, enabling strong performance in long-term time series forecasting. However, these models usually work in a deterministic setting, which does not allow the quantification of the uncertainty of their predictions. Thus, we propose the new Variational Augmented Invertible Koopman AutoEncoder (VAIKAE), in which the latent embedding follows a Gaussian distribution instead of being deterministic. A key property of the VAIKAE architecture is that it leverages normalizing flow models, enabling the use of likelihood computations in the state space of dynamical systems for training a model. We further propose new strategies for uncertainty-aware latent data assimilation with a trained VAIKAE model. The effectiveness of our methods is demonstrated in a series of experiments on long-term time series forecasting benchmarks.

\end{abstract}

\section{Introduction}





Efficient dynamical models are necessary to represent systems whose governing equations are either unknown~\citep{ljung2010perspectives} or too costly to integrate numerically at the required resolution~\citep{kochkov2021machine}. Neural networks~\citep{legaard2023constructing} and the Koopman operator theory~\citep{brunton2022modern} have both recently gained popularity for modeling dynamical systems and are often used jointly~\citep{lusch2018deep}, notably under the framework of Koopman autoencoders (KAEs,~\cite{nayak2025temporally}). KAEs seek a projection from the state space using an autoencoder such that the latent dynamics are approximately linear. Established linear theory, including efficient long-horizon predictions, makes KAEs attractive for modeling dynamical systems. However, most of the existing KAE models are deterministic, and we argue that modeling dynamical systems in the presence of observation and process noise calls for a stochastic formulation.

We consider a discrete autonomous dynamical system with an unknown one-time-step evolution function $\mathcal{M}: \mathbb{R}^n \to \mathbb{R}^n$. One can seek to characterize such a dynamical system by using only some available observations of its state over time. Although many other configurations are possible, we will assume that observations $\mathbf{y}_t \in \mathbb{R}^n$ are full but noisy measurements of the state $\mathbf{x}_t \in \mathbb{R}^n$, yielding 
\begin{align}
    \label{eq:state_equation}
    \forall t \in \mathbb{N}, \quad \mathbf{x}_{t+1} &= \mathcal{M}(\mathbf{x}_t) + \boldsymbol{\epsilon}_t, \quad \boldsymbol{\epsilon}_t \sim \mathcal{N}(\mathbf{0}, \boldsymbol{\Sigma_\epsilon}), \\
    \forall t \in [t_0, ..., t_T], \quad \mathbf{y}_t &= \mathbf{x}_t + \boldsymbol{\eta}_t, \quad \boldsymbol{\eta}_t \sim \mathcal{N}(\mathbf{0}, \boldsymbol{\Sigma_\eta}),
    \label{eq:observation_equation}
\end{align}
where $\boldsymbol{\epsilon}_t$ and $\boldsymbol{\eta}_t$ are, respectively, process and observation noise vectors, which are generally assumed to follow zero-mean Gaussian distributions with time-independent covariance matrices $\boldsymbol{\Sigma_\epsilon}$ and $\boldsymbol{\Sigma_\eta}$. We assume that observations are not available for every time index but only for a specific set $[t_0, ..., t_T]$.

Building an approximation $\hat{\mathcal{M}}$ of the unknown state dynamics $\mathcal{M}$ using an observation set $(\mathbf{y}_{t_0}, ..., \mathbf{y}_{t_T})$ is a ubiquitous problem in machine learning~\citep{raissi2019physics,li2021fourier} and Koopman operator applications~\citep{brunton2022modern}, known as system identification. It enables prediction of the evolution of the state $\mathbf{x}_t$ from any given initial condition. Having access to a surrogate $\hat{\mathcal{M}}$ also facilitates data assimilation~\citep{carrassi2018data}, which consists of combining a dynamical model with an observation set to estimate the posterior distribution of the state over time, i.e. $P(\mathbf{x}_t|\mathbf{y}_{t_0}, ...,\mathbf{y}_{t_T})$. In many cases, the observations $\mathbf{y}_t$ are assumed to be noiseless, i.e. $\boldsymbol{\eta}_t = 0$ in~\eqref{eq:observation_equation}.  Many existing system identification methods seek a deterministic approximation of~\eqref{eq:state_equation}, yet this problem has two sources of uncertainty: first, the state dynamics is inherently noisy (unless $\boldsymbol{\epsilon}_t=0$), and thus even a perfect deterministic estimate of the dynamics $\hat{\mathcal{M}} = \mathcal{M}$ does not fully characterize the evolution of the state $\mathbf{x}_t$. This corresponds to the aleatoric uncertainty. In addition, the performance of system identification can be limited by other factors such as the quality and quantity of the available observations as well as the capacity of the model used to compute $\hat{\mathcal{M}}$. This corresponds to the epistemic uncertainty. A thorough discussion of the sources of uncertainty can be found in~\cite{haynes2023creating}. In particular, we have used the original mathematical definition of aleatoric and epistemic uncertainties, but the definitions may vary in the machine learning community.
Stochastic models can estimate uncertainties and address additional tasks such as anomaly detection~\citep{pang2021deep} and change point detection~\citep{truong2020selective}. By fully approximating~\eqref{eq:state_equation}, they also facilitate uncertainty-aware data assimilation. Ideally, these models should be well calibrated, meaning their estimated uncertainties should match their error statistics, as can be assessed synthetically with the spread-skill ratio or more comprehensively with a spread-skill plot~\citep{haynes2023creating}. 

The rest of this manuscript is organized as follows: in section~\ref{sec:background}, we review related work in  Koopman operator theory, probabilistic time series forecasting and data assimilation. In section~\ref{sec:methods}, we present VAIKAE: a new stochastic KAE architecture which, to our knowledge, is the first to enable explicit state likelihood computations. We then outline strategies for training a model and using it for uncertainty-aware data assimilation. In section~\ref{sec:ltsf}, we present the results of VAIKAE on a time series forecasting benchmark where the input is a complete historical time series. In section~\ref{sec:satellite_experiments}, we experiment on a satellite image time series benchmark where the input observations are irregularly sampled in time, and showcase the effectiveness and good calibration of our data assimilation methods in this setting. Section~\ref{sec:conclusion} concludes our work.

\section{Background}
\label{sec:background}

\subsection{Koopman operator theory}
\label{sec:background_koopman}

The Koopman operator theory, first described by~\cite{koopman1931hamiltonian}, has been extensively used in the last few decades for data-driven analysis of nonlinear dynamical systems, following the work of~\cite{mezic2005spectral}. It states that any nonlinear dynamical system can be described by a linear operator acting on its measurement functions, which is infinite-dimensional in the general case. Concretely, we consider a discrete dynamical operator $\mathcal{M}: \mathbb{R}^n \to \mathbb{R}^n$ following~\eqref{eq:state_equation} with no process noise, i.e. $\boldsymbol{\epsilon}_t = 0$. The Koopman operator $\mathcal{K}$ of $\mathcal{M}$ is such that, for any measurement function $g: \mathbb{R}^n \to \mathbb{R}$ and for any state $\mathbf{x}_t \in \mathbb{R}^n$ at an arbitrary time $t$,
\begin{equation}
    \mathcal{K}g(\mathbf{x}_t) \triangleq g \circ \mathcal{M} (\mathbf{x}_t) = g(\mathbf{x}_{t+1}).
\end{equation}
Since this is true for any input state $\mathbf{x}_t$, one can simply write $\mathcal{K}g = g \circ \mathcal{M}$. 
The Koopman operator is linear but difficult to define for nonlinear dynamics $\mathcal{M}$ due to the infinite dimensionality of its input function space.
Many recent methods seek to find finite-dimensional approximations of the Koopman operator $\mathcal{K}$ to model $\mathcal{M}$. In a general framework, such an approximation can be characterized by three components: a square matrix $\mathbf{K} \in \mathbb{R}^{d \times d}$, an embedding function $\Phi: \mathbb{R}^n \to \mathbb{R}^d$ and a decoding function $\psi: \mathbb{R}^d \to \mathbb{R}^n$. $\Phi$ embeds a state $\mathbf{x}_t \in \mathbb{R}^n$ expressed in the natural basis of the dynamical system using a set of $d$ measurement functions, yielding a vector $\mathbf{z}_t = \Phi(\mathbf{x}_t) \in \mathbb{R}^d$. This vector is multiplied by $\mathbf{K}$ in order to get the latent embedding $\mathbf{z}_{t+1}$, which can be decoded back to the state space using $\psi$. This process is summarized by
\begin{equation}
\label{eq:koopman_pred}
    \mathbf{x}_{t+\tau} \approx \hat{\mathbf{x}}_{t+\tau} = \psi(\mathbf{K}^\tau\Phi(\mathbf{x}_t))
\end{equation}
for any chosen prediction time $\tau > 0$. Formally, $\mathbf{K}$ approximates the restriction of the infinite-dimensional Koopman operator $\mathcal{K}$ on the set of $d$ measurement functions represented by $\Phi$. Thus, a fundamental assumption of this approach is that $\Phi$ (approximately) defines a Koopman invariant subspace~\citep{brunton2016koopman}, i.e. a set of measurement functions that is stable by application of the Koopman operator.

Many practical approaches have been proposed for obtaining $\Phi$, as reviewed by~\cite{brunton2022modern}. An early and popular method is dynamic mode decomposition (DMD,~\cite{schmid2010dynamic}), which defines~$\Phi$ and~$\psi$ as identity functions. This means assuming that the set of natural measurement functions consisting of projections of $\mathbf{x} \in \mathbb{R}^n$ to its $n$ variables is approximately Koopman invariant, i.e. that the dynamical system under study is approximately linear. The matrix $\mathbf{K}$ is then estimated from a dataset of consecutive system states.
Extended dynamic mode decomposition (eDMD,~\cite{williams2015data}) generalizes DMD by using a hand-designed~$\Phi$ that includes the natural measurement functions. $\psi$ is then obtained by projecting the latent embedding onto its~$n$ first variables. eDMD converges to the Koopman operator as its latent embedding size~$d$ grows to infinity~\citep{korda2018convergence}, yet obtaining good practical performance often requires physical insight on the studied dynamical system and a large latent dimension~$d$. Thus, many subsequent works have used neural autoencoders for learning $\Phi$ and $\psi$ as encoding and decoding functions.

The Koopman autoencoder (KAE) models generally define $\Phi$, $\psi$ and $\mathbf{K}$ as three learnable components. 
We distinguish two desirable properties for such models:
\begin{itemize}
    \item The learned embedding $\Phi$ should be as close to Koopman invariant as possible, i.e. $\mathbf{K}\Phi(\mathbf{x}_t) \approx \Phi(\mathbf{x}_{t+1})$. This criterion is linked to the expressivity of $\Phi$ and to the size $d$ of the latent embedding.
    \item The learned embedding should be as close to invertible as possible, i.e. $\psi \circ \Phi(\mathbf{x}_t) \approx \mathbf{x}_t$. Ideally, this reconstruction should be analytically exact.
\end{itemize}
A model that perfectly respects these two properties would perfectly represent the state dynamics. Early KAE models~\citep{lusch2018deep,otto2019linearly,Li2020Learning,azencot2020forecasting} train two neural networks for $\Phi$ and $\psi$ with several loss terms including a reconstruction loss so that~$\psi \circ \Phi$ is close to the identity function. This should satisfy the first desired criterion if the model for $\Phi$ is expressive enough and uses a large latent size $d$. But it inevitably leads to a reconstruction error with the composition $\psi \circ \Phi$. 
Thus, more recent works~\citep{meng2024koopman,jin2024extended,hou2024invertible} use analytically invertible neural architectures, such as normalizing flows~\citep{kobyzev2020normalizing}, so that $\psi = \Phi^{-1}$, ensuring an exact reconstruction of the input state from its embedding. A drawback of this choice is that the learned embedding must match the state dimension, i.e. $d=n$, whereas a larger embedding is beneficial for the first criterion of finding a Koopman invariant subspace. To address this issue, \cite{meng2024koopman,jin2024extended} pad the embedding with zeros, yet this approach still lacks expressivity. Instead, \cite{frion2025augmented,lupascu2026predicting} augment the learned invertible embedding with a second encoder that has no invertibility constraint. This enables greater representational power to learn a Koopman invariant subspace while still guaranteeing an exact reconstruction of the input state. However, all methods mentioned so far are restricted to deterministic prediction, and thus we now turn our attention to probabilistic methods.

\subsection{Stochastic time series forecasting}
\label{sec:background_stochastic}

Stochastic rather than deterministic models can capture complex conditional or unconditional probability distributions~\citep{sengar2025generative} or quantify predictive uncertainty~\citep{haynes2023creating}. Ensemble averages also improve deterministic metrics such as root mean squared error over a single prediction~\citep{milinski2020large}. In practice, ensembling extended the horizon of skillful predictions for chaotic systems such as the global atmosphere~\citep{nathaniel2024chaosbench}, which lead to a large shift towards stochastic neural models~\citep{oskarsson2024probabilistic,price2025probabilistic,alet2025skillful,lang2026aifs,agarwal2026skillful}. Simple and successful techniques to obtain stochasticity with generic neural network architectures include Monte Carlo dropout~\citep{gal2016dropout}, stochastic weight averaging~\citep{izmailov2018averaging}, Bayesian neural networks~\citep{jospin2022hands} and model ensembling~\citep{lakshminarayanan2017simple,frion2024koopman}. 

Diffusion models~\citep{yang2023diffusion} are gaining increasing popularity for probabilistic time series forecasting~\citep{liao2026deep}. TimeGrad~\citep{rasul2021autoregressive} leverages a diffusion model that estimates the distribution of the state at each time step based on the hidden state of a recurrent neural network. CSDI~\citep{tashiro2021csdi} uses a score-based diffusion model conditioned on observed data, and can solve multiple tasks including time series imputation and long-term forecasting. TimeDiff~\citep{shen2023non} builds a conditioning signal that combines future mixup (inspired by the mixup from~\cite{zhang2018mixup}) and autoregressive initialization. TMDM~\citep{li2024transformer} builds on the NSFormer architecture~\citep{liu2022non} and minimizes an evidence lower bound (ELBO) loss to estimate the posterior distribution of the time series given an input history.  The authors of $\mathrm{D^3U}$~\citep{li2025diffusion} first train a deterministic model to learn the conditional mean, and then train a DDPM~\citep{ho2020denoising} to model the probabilistic part of the prediction.

Another line of work relies on the Koopman operator theory. The authors of~\cite{pan2020physics} implement the trainable components of a KAE architecture as Bayesian neural networks~\citep{jospin2022hands}, enabling probabilistic outputs. DeSKO~\citep{han2022desko} uses an encoder that outputs the mean and variance of a latent diagonal Gaussian distribution 
for uncertainty-aware model predictive control. KoVAE~\citep{naiman2024generative} is a variational autoencoder model that projects each latent sequence to its best linear fit using DMD. Deep Probabilistic Koopman~\citep{mallen2024deep} models probability distributions from which the parameters follow a quasi-periodic evolution in time. KooNPro~\citep{zheng2025koonpro}, inspired by~\cite {lusch2018deep}, designs a KAE with an auxiliary model that estimates a Gaussian probability distribution for the spectrum of the latent dynamics, and derives an associated ELBO criterion~\citep{garnelo2018conditional}.


\subsection{Data assimilation with automatic differentiation and neural networks}
\label{sec:background_da}

Data assimilation~\citep{carrassi2018data} is a Bayesian framework that leverages both a dynamical model $\mathcal{M}$ and a set of imperfect observations $\mathbf{y}_t$ to reconstruct the full state of a system $\mathbf{x}_t$ over time. 
Its joint use with machine learning is a rich and ongoing field of study~\citep{bocquet2023surrogate,cheng2023machine}. 
In particular, variational data assimilation, consisting of gradient descent on a Bayesian maximum a posteriori cost, classically requires the complex hand-derivation of an adjoint model of the dynamics. This derivation can be avoided by implementing the dynamics in an automatic differentiation framework such as PyTorch~\citep{paszke2017automatic} or JAX~\citep{jax2018github}, as discussed by e.g.~\cite{gelbrecht2023differentiable,sapienza2024differentiable,frion2026adda}. Alternatively, one can use a neural emulator that approximates the true dynamics and is differentiable by design~\citep{nonnenmacher2021deep,hatfield2021building}. 

Following this second approach, latent data assimilation consists of solving a data assimilation problem in the latent space of a trained neural emulator. It can rely on various classical data assimilation methods, e.g. ensemble-based~\citep{peyron2021latent} or variational~\citep{melinc20243d} methods. The motivations for resorting to latent data assimilation include working in a lower-dimensional space than the physical state space to reduce the computational cost~\citep{peyron2021latent,melinc20243d} and building a non-Gaussian prior distribution~\citep{pasmans2026ensemble,fan2026physically}. Most related to the present work, some methods~\citep{frion2024neural,shoji2025data,frion2025augmented,tong2026latent} perform data assimilation in the latent space of a KAE model to benefit from linear latent dynamics.

\section{Proposed methods}
\label{sec:methods}

\subsection{A new Koopman autoencoder model architecture: VAIKAE}
\label{sec:methods_VAIKAE}

We base our new KAE model on the recently proposed Augmented Invertible Koopman AutoEncoder (AIKAE,~\cite{frion2025augmented}). A visual representation of the AIKAE architecture is shown in appendix~\ref{sec:AIKAE}. It comprises 3 learnable components: an invertible encoder $\phi: \mathbb{R}^{n} \to \mathbb{R}^n$ (implemented as a normalizing flow), a Koopman matrix $\mathbf{K} \in \mathbb{R}^{d \times d}$ and an augmentation encoder $\chi: \mathbb{R}^n \to \mathbb{R}^p$. In this model, the latent Koopman embedding $\Phi(\mathbf{x}_t)$ of $\mathbf{x}_t \in \mathbb{R}^n$ is defined as the concatenation of $\mathbf{z}_t^i = \phi(\mathbf{x}_t)$ and $\mathbf{z}_t^a = \chi(\mathbf{x}_t)$, so that the model has an unrestricted latent dimension $d=n+p$ while still exactly reconstructing the input state using the analytical inverse $\phi^{-1}$ of $\phi$. We respectively use superscripts~$\cdot^i$ and~$\cdot^a$ to represent the invertible (first~$n$ components) and augmentation (last~$p$ components) parts of a vector in~$\mathbb{R}^d$.

\begin{figure}
\begin{center}
    \includegraphics[width=15cm]{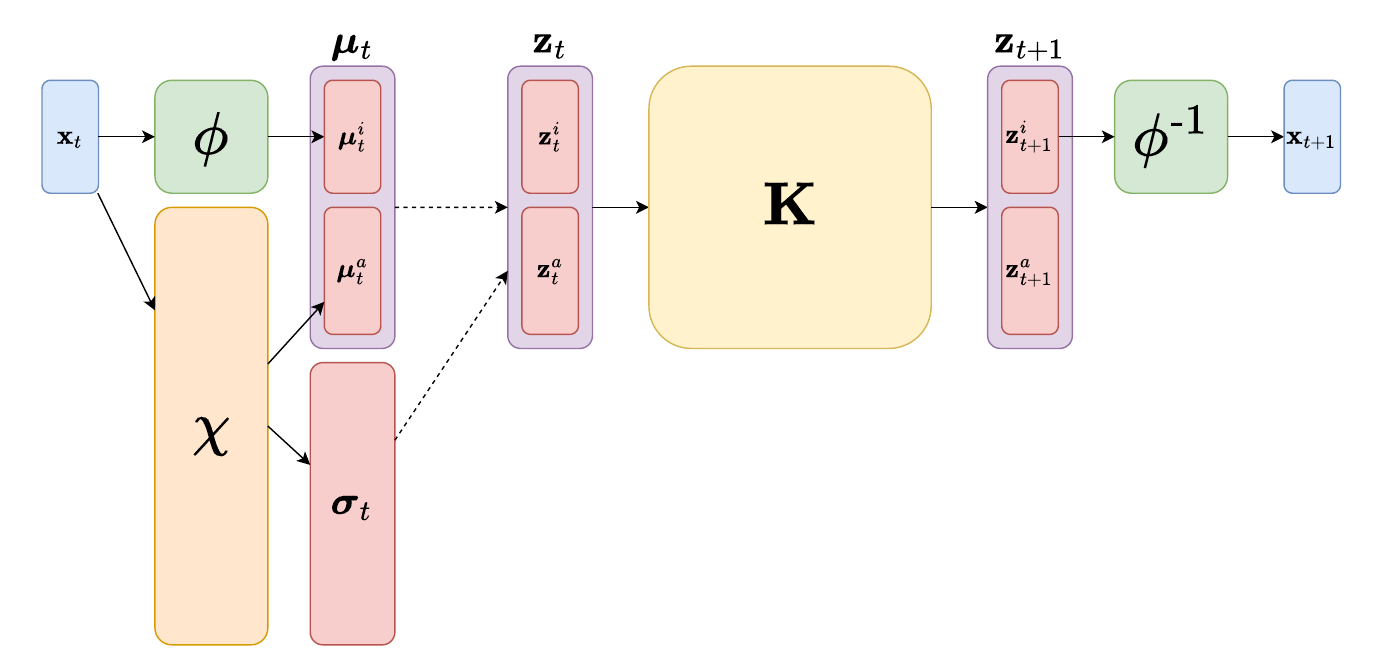}
\end{center}
\caption{Graphical representation of the VAIKAE architecture. The two dashed lines represent sampling from a diagonal Gaussian distribution, while the full lines represent deterministic operations.}
\label{fig:VAIKAE}
\end{figure}

Here, we extend the AIKAE framework to a Variational Augmented Invertible Koopman AutoEncoder (VAIKAE), which substitutes the deterministic latent embedding of the AIKAE with a probabilistic one. Concretely, the embedding of a state $\mathbf{x}_t \in \mathbb{R}^n$ by the VAIKAE is a diagonal Gaussian distribution\footnote{The choice of a diagonal Gaussian distribution in the latent space is common for generative models (e.g.~\cite{kingma2013auto,dinh2017density,yang2023diffusion,naiman2024generative}) due to its flexibility and low parameter count.} defined by its mean $\boldsymbol{\mu}_t \in \mathbb{R}^d$ and (diagonal) covariance $\boldsymbol{\sigma}_t \in \mathbb{R}^d$. These mean and covariance are obtained by repurposing the augmentation encoder $\chi$ so that it additionally outputs the variance coefficients, leading to:
\begin{equation}
    \chi(\mathbf{x}_t) = \begin{pmatrix}
        \boldsymbol{\mu}_t^a \\
        \boldsymbol{\sigma}_t
    \end{pmatrix}.
\end{equation}
Thus, the output of $\chi: \mathbb{R}^n \to \mathbb{R}^{p+d}$ is decomposed into 2 parts: the first $p$ components represent the mean $\boldsymbol{\mu}_t^a$ of the augmentation part of the encoding and the last $d$ components represent the diagonal covariance of the global latent embedding. The invertible encoder $\phi$ keeps an analogous role as in AIKAE, here outputting $\boldsymbol{\mu}_t^i = \phi(\mathbf{x}_t)$. The VAIKAE architecture is summarized in figure~\ref{fig:VAIKAE}.

Another design choice would have been to define a third encoder to estimate the latent diagonal variance coefficients $\boldsymbol{\sigma}_t$ while $\phi$ and $\chi$ produce the means of the invertible and augmentation parts of the latent embedding. However, we instead let $\chi$ additionally learn the variance coefficients, thus sharing weights with the mean $\boldsymbol{\mu}_t^a$ of the augmentation encoding. One advantage of this is that VAIKAE requires only few additional parameters (all in the output layer of $\chi$) relative to its deterministic AIKAE counterpart.

\subsection{Training a VAIKAE}
\label{sec:training}

In this section, we show that the VAIKAE framework enables us to explicitly evaluate the predicted probability density of a state value given an earlier observed value. This mathematical derivation is then used to design a training criterion for obtaining calibrated stochastic predictions with the model.

We work with the assumptions of equations~\ref{eq:state_equation} and~\ref{eq:observation_equation}, where $\mathbf{y}_t \in \mathbb{R}^n$ is a noisy observation of the true state $\mathbf{x}_t \in \mathbb{R}^n$ at time $t$. Thus, assuming here that we only have access to $\mathbf{y}_0$ at time $0$, our objective is to characterize the posterior probability distribution $P_{\mathbf{x}_{t}}(\cdot|\mathbf{y}_0)$, for any prediction time $t \geq 0$. 
We start from the initial latent embedding's probability distribution $P_{\mathbf{z}_0}(\cdot|\mathbf{y}_0) \sim \mathcal{N}(\boldsymbol{\mu}_0, \boldsymbol{\Sigma}_0 = \text{diag}(\boldsymbol{\sigma}_0))$, where the conditional mean $\boldsymbol{\mu}_0$ and variance coefficients $\boldsymbol{\sigma}_0$ are respectively obtained with
\begin{align}
    &\boldsymbol{\mu}_0 = \begin{pmatrix}
        \phi(\mathbf{y}_0) \\
        \chi(\mathbf{y}_0)_{1:p}
    \end{pmatrix},
    \label{eq:latent_mean}
    \\
    &\boldsymbol{\sigma}_0 = \chi(\mathbf{y}_0)_{p+1:p+d}
    \label{eq:latent_var}.
\end{align}
From here on, all subsequent latent embeddings $\mathbf{z}_{t}$ are linearly related to $\mathbf{z}_0$ through $\mathbf{z}_{t} = \mathbf{K}^t \mathbf{z}_0$. Thus, when conditioned on $\mathbf{y}_0$, they also follow Gaussian distributions as:
\begin{align}
    \label{eq:latent_cond_distrib}
    & P_{\mathbf{z}_t}(\cdot|\mathbf{y}_0) \sim \mathcal{N}(\boldsymbol{\mu}_t, \boldsymbol{\Sigma}_t) 
\end{align}
with mean $\boldsymbol{\mu}_t = \mathbf{K}^t \boldsymbol{\mu}_0$, covariance $\boldsymbol{\Sigma}_t = \mathbf{K}^t\boldsymbol{\Sigma}_0(\mathbf{K}^t)^\intercal$, and $\cdot^\intercal$ denoting matrix transposition. Importantly, while $\boldsymbol{\Sigma}_0$ is a diagonal matrix, $\boldsymbol{\Sigma}_t$ is likely to be a full covariance matrix when $\mathbf{K}$ is not diagonal. 

From this point, one can use the properties of the normalizing flow $\phi$ to evaluate the probability density function of $\mathbf{x}_t$ given the distribution of $\mathbf{z}_t = \phi(\mathbf{x}_t)$, and ultimately evaluate the likelihood given the observed value $\mathbf{y}_0$. We apply the following change of variable formula (discussed in e.g.~\cite{dinh2014nice}):
\begin{equation}
\label{eq:change_of_variable}
    P_{\mathbf{x}_t}(\mathbf{x}) = P_{\mathbf{z}_t^i}(\mathbf{\phi(\mathbf{x})}) \left|\det\frac{\partial\phi(\mathbf{x})}{\partial \mathbf{x}} \right|. 
\end{equation}
Concerning the first factor $P_{\mathbf{z}_t^i}(\mathbf{\phi(\mathbf{x})})$, it should be noted that $\mathbf{z}_t^i$ is a marginal distribution of $\mathbf{z}_t$, and thus $\mathbf{z}_{t}^i$ is also Gaussian when $\mathbf{z}_t$ is Gaussian~\citep{bishop2006pattern}. Its moments can be obtained by taking the first $n$ components $\boldsymbol{\mu}_t^i$ of the mean and the upper-left $n \times n$ block $\boldsymbol{\Sigma}_t^i$ of the covariance of $\mathbf{z}_t$.
Besides, the determinant of the Jacobian matrix $\frac{\partial\phi(\mathbf{x})}{\partial \mathbf{x}}$ is not easy to obtain in general, yet $\phi$ is here a normalizing flow model, and is therefore specifically designed to have a tractable and easily computable Jacobian. 
When conditioning~\eqref{eq:change_of_variable} on an observed value of $\mathbf{y}_0$ and injecting~\eqref{eq:latent_cond_distrib}, we obtain:
\begin{equation}
\label{eq:state_likelihood}
    P_{\mathbf{x}_t}(\mathbf{x} | \mathbf{y}_0) = 
    P_{\mathbf{z}_t^i}(\phi(\mathbf{x}) | \mathbf{y}_0)\left|\det\frac{\partial\phi(\mathbf{x})}{\partial \mathbf{x}} \right| 
    = \mathcal{N}(\phi(\mathbf{x}); \boldsymbol{\mu}_t^i, \boldsymbol{\Sigma}_t^i) \left|\det\frac{\partial\phi(\mathbf{x})}{\partial \mathbf{x}} \right|.
\end{equation}
As a practical training criterion, one can use this equation to maximize the likelihood $P_{\mathbf{x}_t}(\mathbf{y}_t|\mathbf{y}_0)$ of subsequent observations $\mathbf{y}_t$ when predicting from an input observation $\mathbf{y}_0$. Importantly, this strategy only accounts for the marginal distributions on the state variables at each time step, and thus does not ensure coherent predicted trajectories. Yet, we can still obtain temporally coherent trajectories by drawing samples from $P_{\mathbf{z}_0}(\cdot|\mathbf{y}_0)$ and then deterministically propagating them to generate one trajectory from each of these samples, rather than independently drawing samples from the marginal distributions at each time step. 

We now describe a complete strategy for training the VAIKAE model. Let us assume, for simplicity, that the training dataset is composed of $N$ sets of observations $\mathbf{Y}_1, ..., \mathbf{Y}_N$ such that, for any $1 \leq i \leq N$, $\mathbf{Y}_i = (\mathbf{y}_{i,t_{i,0}}, ..., \mathbf{y}_{i,t_{i,T_i}})$ with $0 = t_{i,0} < ... < t_{i,T_i}$ representing the indices where observations are available. The individual sets of observations are usually overlapping slices of longer sets of observations. We define $\theta$ as the concatenation of the coefficients of $\mathbf{K}$ and the trainable parameters of $\phi$ and $\chi$. We take inspiration from the loss function of the deterministic AIKAE model to design three loss function terms:
\begin{itemize}
    \item A prediction loss $L_{pred}(\theta) = \sum_{i=1}^N \sum_{\tau=0}^{T_i} \mathbb{E}_{\mathbf{x} \sim P_{\mathbf{x}_{i,t_{i,\tau}}}(\cdot | \mathbf{y}_{i,0})} ||\mathbf{x} - \mathbf{y}_{i,t_{i,\tau}}||^2_2$ consisting in the mean squared error between sampled predictions obtained from~\eqref{eq:koopman_pred} and the corresponding true states.
    \item A linearity loss $L_{lin}(\theta) = \sum_{i=1}^N \sum_{\tau=0}^{T_i} \mathbb{E}_{\mathbf{z} \sim P_{\mathbf{z}_{i,t_{i,\tau}}}(\cdot | \mathbf{y}_{i,0})} ||\mathbf{z} - \phi(\mathbf{y}_{i,t_{i,\tau}})||^2_2$, which is meant to ensure that the latent embeddings of observations of the same state over time are truly linearly related.
    \item An orthogonality loss $L_{orth}(\theta) = ||\mathbf{K}\mathbf{K}^\intercal - \mathbf{I}_d||^2_2$, which ensures that the eigenvalues of the learned $\mathbf{K}$ remain close to the unit circle to obtain stable dynamics.
\end{itemize}
These loss terms are studied in a deterministic setting in~\cite{frion2024neural}. While~$L_{orth}$ was shown to promote long-term stability, $\mathbf{K}$ could also be constrainted to be unitary by construction~\citep{zhang2024learning}. 

In practice, the expected values in $L_{pred}$ and $L_{lin}$ are approximated by sampling from the latent Gaussian distribution $P_{\mathbf{z}_{i,0}}(\cdot | \mathbf{y}_{i,0})$ and then advancing these samples deterministically to obtain samples from all relevant variables. However, using a loss function based only on these terms would result in a collapse of the learned variance vectors $\boldsymbol{\sigma}_0$ to $0$, reducing to deterministic predictions. To force some stochasticity into the model, we add the new likelihood loss function $L_{lkl}$, which is directly derived from~\eqref{eq:state_likelihood} as:
\begin{equation}
\label{eq:complete_loss}
    L_{lkl}(\theta) = \sum_{i=1}^N \sum_{\tau=0}^{T_i} - \log (P_{\mathbf{x}_{i,t_{i,\tau}}}(\mathbf{y}_{i,t_{i,\tau}} | \mathbf{y}_{i,0})).
\end{equation}
Since the negative log-likelihood is a proper scoring rule (see appendix section~\ref{sec:proper_scoring_rules}), it should favor well-calibrated predictions.
Finally, one can now construct the global loss function for VAIKAE:
\begin{equation}
    L(\theta) = L_{pred}(\theta) + \alpha L_{lin}(\theta) + \beta L_{orth}(\theta) + \gamma L_{lkl}(\theta),
\end{equation}
where $\alpha, \beta, \gamma$ are relative weights. Depending on the application, one may set $\alpha = 0$ and/or $\beta = 0$, yet $L_{pred}$ is always present as the main loss term, while $L_{lkl}$ is necessary to obtain stochastic predictions in practice.

\subsection{Performing data assimilation in the latent space of a trained VAIKAE}
\label{sec:methods_assimilation}

Here, we consider the problem of using multiple observations $\mathbf{y}_t$ to make stochastic predictions, 
in the data assimilation context from section~\ref{sec:background_da}. Given an observations set $(\mathbf{y}_{t_0}, ..., \mathbf{y}_{t_T})$, \cite{frion2025augmented} solves a strong-constraint 4D-Var problem in the latent space of an AIKAE model, which can be written as
\begin{equation}
    \mathbf{z}_* = \underset{\mathbf{z}_0 \in \mathbb{R}^d}{\textrm{arg min}} \sum_{\tau=0}^T ||\phi^{-1}(\mathbf{K}^{t_\tau}\mathbf{z}_0) - \mathbf{y}_{t_\tau}||^2 \label{eq:LDA_deterministic},
\end{equation}
$\mathbf{z}_*$ can then be used to produce predictions at any time $t \geq t_0$. 
Equation~\ref{eq:LDA_deterministic} can be conveniently solved with automatic differentiation. Crucially, one can use $\Phi(\mathbf{y}_{t_0})$ as the initial guess that solves~\eqref{eq:LDA_deterministic}. Due to the general non-convex nature of the problem, this initialization both reduces the number of gradient steps needed for convergence and improves the quality of the end result.

In a VAIKAE, $\Phi$ predicts not only a single value of $\mathbf{z}_0$ but the mean and variance of a latent diagonal Gaussian distribution. Equation~\ref{eq:LDA_deterministic} can thus be adapted for stochastic predictions by solving 4D-Var on the mean and then decoding this mean to estimate an associated uncertainty with~$\chi$. This can be written as:
\begin{align}
    &\boldsymbol{\mu}_* = \underset{\boldsymbol{\mu}_0 \in \mathbb{R}^d}{\text{arg min}} \sum_{\tau=0}^T ||\phi^{-1}(\mathbf{K}^{t_\tau}\boldsymbol{\mu}_0) - \mathbf{y}_{t_\tau}||^2 \label{eq:LDA_method1_eq1}, \\
    &\boldsymbol{\sigma}_* = \chi(\phi^{-1}(\boldsymbol{\mu}_*^i))_{p+1:p+d}. \label{eq:LDA_method1_eq2}
\end{align}
As a reminder, $\boldsymbol{\mu}_*^i$ is the invertible part of $\boldsymbol{\mu}_*$, i.e. its first $n$ components. This method can provide uncertainty quantification, yet it will likely be under-confident since using multiple observations should significantly improve the skill of the prediction with regard to predictions from a single observation, while retaining the same spread.
Alternatively, one can reformulate the original formulation of 4D-Var to solve for the parameters of a Gaussian distribution instead of a point estimate. We propose to minimize the continuous ranked probability score (CRPS), which is a proper scoring rule and performs strongly on fitting forecasting ensembles: see e.g.~\cite{gneiting2007strictly} and appendix~\ref{sec:proper_scoring_rules}. Concretely, we solve: 
\begin{equation}
\label{eq:LDA_method2}
    \boldsymbol{\mu}_*, \boldsymbol{\sigma}_* = \underset{\boldsymbol{\mu}_0, \boldsymbol{\sigma}_0}{\text{arg min}} \; \sum_{\tau=0}^T \mathrm{CRPS}(\mathbf{x}_{t_\tau}, \mathbf{y}_{t_\tau}).
\end{equation}
Note that, in practice, we perform each gradient descent step by drawing multiple Monte Carlo samples from $\mathbf{z}_0 \sim \mathcal{N}(\boldsymbol{\mu}_0, \mathrm{diag(\boldsymbol{\sigma}_0}))$ with the reparameterization trick from~\cite{kingma2013auto} and advancing each of them deterministically to obtain samples from $(\mathbf{x}_{t_0}, ..., \mathbf{x}_{t_T})$ conditioned on $\mathbf{z}_0$. We then use the sample-based computation of the CRPS with~\eqref{eq:crps}. As detailed in appendix~\ref{sec:proper_scoring_rules}, we average the CRPS obtained for all individual variables of the state. 
Analogously to the AIKAE-based assimilation in~\eqref{eq:LDA_deterministic}, one can use the pre-trained VAIKAE to obtain initial guesses for both the initial mean and variance in~\eqref{eq:LDA_method2}, which enables a faster convergence and a better end result. 
While minimizing the cost of~\eqref{eq:LDA_method2} means that we no longer benefit from the rigorous Bayesian maximum a posteriori formalism of 4D-Var, we will show in section~\ref{sec:satellite_DA} that it still enables us to obtain well-calibrated posterior distribution estimates.

\section{Probabilistic long-term time series forecasting}
\label{sec:ltsf}

Here, we test the performance of our VAIKAE model on the popular "Informer benchmark", named after a method that popularized it~\citep{zhou2021informer}. This benchmark consists of a set of time series datasets, from which one usually extracts a set of $L$ consecutive time steps and uses it as an input to predict the state over the following $T$ time steps. While many methods leverage the joint information of all variables in the input time series~\citep{zhou2021informer, zhou2022fedformer, zhang2023crossformer,liu2024itransformer}, some recent methods have obtained strong performance by considering the variables of the time series independently from each other~\citep{zeng2023transformers, nie2023a, frion2025augmented}, and we adopt this second approach.

Although multiple methods (e.g.~\cite{zhou2022fedformer,zhang2023crossformer,zeng2023transformers,nie2023a,liu2024itransformer,frion2025augmented}) have been proposed to perform long-term forecasting on this benchmark in a deterministic setup, evaluated solely through their mean squared error (MSE) and mean average error (MAE), we here focus on probabilistic models, which are evaluated with additional metrics such as the CRPS. A general presentation of our metrics can be found in appendix~\ref{sec:metrics}.
%


\begin{table}[tb]
    \caption{Summary of probabilistic long-term time series forecasting results. For each dataset and metric, the best result is in \textbf{bold} and the second best result is \underline{underlined}. 
    }
    \vspace{1mm}
    \label{tab:benchmark}
    \centering
    \scalebox{0.9}
    {
    \begin{tabular}{|c|c|c|c|c|c|c|c|}
    \hline
        \multicolumn{2}{|c|}{$\mathbf{Model}$} & $\mathbf{VAIKAE}$ & $\mathbf{D^3U}$ & $\mathbf{TMDM}$ & $\mathbf{TimeDiff}$ &$\mathbf{CSDI}$ & $\mathbf{TimeGrad}$ \\
        \hline
        \multirow{3}{1em}{\rotatebox[origin=c]{90}{ETTm1}} & MSE & \underline{0.370} & \textbf{0.363} & 0.607 & 0.796 & 0.867 & 1.716 \\
        & MAE & \textbf{0.385} & \underline{0.386} & 0.558 & 0.577 & 0.690 & 1.057 \\
        & CRPS & \underline{0.299} & \textbf{0.285} & 0.429 & 0.454 & 0.773 & 0.665 \\
        \hline
        \multirow{3}{1em}{\rotatebox[origin=c]{90}{ETTm2}} & MSE & \underline{0.253} & \textbf{0.241} & 0.524 & 0.284 & 1.291 & 1.385 \\
        & MAE & \underline{0.319} & \textbf{0.302} & 0.493 & 0.342 & 0.576 & 0.732 \\
        & CRPS & \underline{0.247} & \textbf{0.243} & 0.380 & 0.316 & 0.625 & 0.785 \\
        \hline
        \multirow{3}{1em}{\rotatebox[origin=c]{90}{Weather}} & MSE & \textbf{0.208} & \underline{0.222} & 0.244 & 0.277 & 0.842 & 0.885 \\
        & MAE & \textbf{0.254} & \underline{0.264} & 0.286 & 0.331 & 0.523 & 0.551 \\
        & CRPS & \textbf{0.199} & \underline{0.207} & 0.226 & 0.293 & 0.508 & 0.482 \\
        \hline
        \multirow{3}{1em}{\rotatebox[origin=c]{90}{Solar}} & MSE & \textbf{0.233} & \underline{0.237} & 0.295 & 1.169 & 0.848 & 1.211 \\
        & MAE & \underline{0.288} & \textbf{0.270} & 0.317 & 0.936 & 0.818 & 1.004 \\
        & CRPS & \underline{0.235} & \textbf{0.186} & 0.375 & 0.900 & 0.649 & 0.783 \\
        \hline
        \multirow{3}{1em}{\rotatebox[origin=c]{90}{ECL}} & MSE & \textbf{0.172} & \underline{0.179} & 0.222 & 0.730 & 0.553 & 0.645 \\
        & MAE & \textbf{0.264} & \underline{0.267} & 0.329 & 0.690 & 0.795 & 0.723 \\
        & CRPS & \textbf{0.199} & \underline{0.202} & 0.446 & 0.475 & 0.465 & 0.503 \\
        \hline
        \multirow{3}{1em}{\rotatebox[origin=c]{90}{Traffic}} & MSE & \textbf{0.452} & \underline{0.468} & 0.721 & 1.465 & 0.921 & 0.932 \\
        & MAE & \underline{0.305} & \textbf{0.299} & 0.411 & 0.851 & 0.678 & 0.807 \\
        & CRPS & \underline{0.243} & \textbf{0.232} & 0.552 & 0.671 & 0.612 & 0.657 \\
        \hline
    \end{tabular}
    }
    
\end{table}

\begin{figure}[tb]
\begin{center}
    \includegraphics[width=15.5cm]{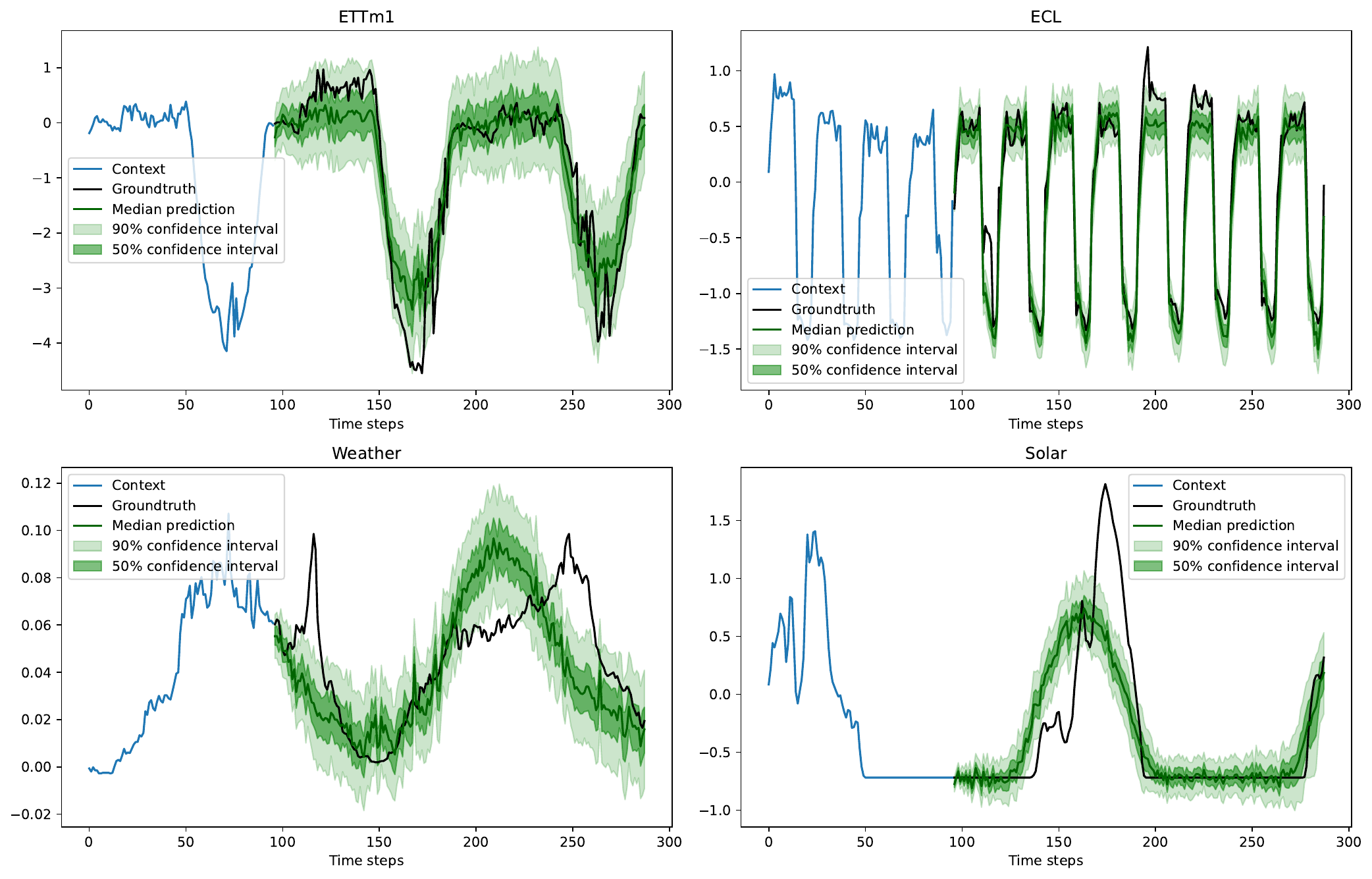}
\end{center}
\caption{Representative examples from the test sets of ETTm1 (top left), ECL (top right), Weather (bottom left) and Solar (bottom right). For each example, we show the 96-step input context, the subsequent 192-step groundtruth and the median VAIKAE prediction with corresponding $50\%$ and $90\%$ confidence intervals.}
\label{fig:Informer_examples}
\end{figure}

We use the setup of~\cite{li2025diffusion}, where the input length is $L=96$ and the prediction length is $T=192$. Similar to~\cite{frion2025augmented}, we use a delay embedding of the input time series, which means that the input size is $n= L=96$. The size of the augmentation encoding is set to $p=32$, so that the global latent space of the model is of size $d=n+p=128$. Further implementation details can be found in appendix~\ref{sec:implementation_Informer}. Our baselines are five recent and competitive methods based on diffusion models: TimeGrad~\citep{rasul2021autoregressive}, CSDI~\citep{tashiro2021csdi}, TimeDiff~\citep{shen2023non}, TMDM~\citep{li2024transformer} and $\mathrm{D^3U}$~\citep{li2025diffusion}. These methods are discussed in section~\ref{sec:background_stochastic}. Their results are directly taken from~\cite{li2025diffusion}.

The MSE, MAE and CRPS of all models, summarized in table~\ref{tab:benchmark}, exhibit strong performance of VAIKAE, which obtains either the best or second best result on all dataset-metric combinations. 
We additionally plot representative examples from the test sets of different datasets in figure~\ref{fig:Informer_examples}. In most of these examples, the $90\%$ confidence interval of the predictions includes the true state during most of the predicted intervals, which corresponds to the expected behavior, and suggests that the uncertainties are well calibrated. An exception is the example from the Solar dataset, where the groundtruth lies outside of the 90\% confidence interval for a large portion of the predicted time steps, which shows that our prediction is overconfident for this dataset. This observation is consistent with the relatively poor CRPS obtained by VAIKAE for Solar in table~\ref{tab:benchmark} compared to the strong associated MSE result.

\section{Data assimilation on satellite image time series}
\label{sec:satellite_experiments}

Here, we test our VAIKAE model on a satellite image time series forecasting benchmark, which was previously considered by e.g.~\cite{frion2024neural,frion2025augmented}. This benchmark consists of multispectral images with 10 separate spectral bands from the visible and infrared spectral domains. The images are obtained from the Sentinel-2 satellite constellation with a time step of 5 days and a spatial resolution of 10 meters. They are irregularly sampled due to the meteorological conditions that prevent obtaining high-quality images of the ground on most dates, which corresponds to the observation setup of~\eqref{eq:observation_equation}. The benchmark considers two areas, of $5 \times 5$ km (i.e. $500 \times 500$ pixels) each, which are respectively named "Fontainebleau" and "Orléans", and both mostly covered by forest, although with differing properties. 

Following~\cite{frion2024neural,frion2025augmented}, we train our VAIKAE model to predict pixelwise dynamics on a restricted time domain of the data from the Fontainebleau area, and then test its performance on both the held out time domain of the Fontainebleau area (i.e. temporal extrapolation) and the full time domain of the Orléans area (i.e. spatial extrapolation). In contrast to~\cite{frion2025augmented}, we here train our model on a larger spatial subdomain, of size $300 \times 300$ pixels (i.e. 9 square kilometers), as shown in figure~\ref{fig:Sentinel_sample}, and we directly use irregularly-sampled data rather than a time-interpolated time series. The considered time domains for training and testing are consistent with prior works, and are respectively composed of $T_{train}=242$ time steps (roughly three and a half years) and the following $T_{test}=100$ time steps (roughly one and a half years).

\subsection{Comparison of VAIKAE with other Koopman-based stochastic models}
\label{sec:Sentinel_training}

We consider several KAE methods for probabilistic forecasting of the pixelwise reflectance vectors:
\begin{itemize}
    \item Our \textbf{VAIKAE} model, trained with the loss function of~\eqref{eq:complete_loss}.
    \item An ablated version of VAIKAE that does not include an augmentation part in its latent embedding, but only an invertible encoding, thus facing the restriction $d=n$. We refer to this model as \textbf{VIKAE}.
    \item Another ablation, named \textbf{VKAE}, where the latent embedding is obtained by a regular autoencoder, similar to~\cite{frion2024neural} yet also with a Gaussian distribution for its latent embedding.
    \item An ensemble of 16 deterministic KAE models, jointly trained with a variance-promoting loss term, following~\cite{frion2024koopman}. We call this approach \textbf{KAE ensemble}.
\end{itemize}

\begin{table}[t]
\centering
\caption[size=9pt]{Forecasting CRPS for different methods and areas}
\label{tab:Sentinel_forecasting}
\begin{tabular}{|c|c|c|}
\hline
 & CRPS on the Fontainebleau area  & CRPS on the Orléans area \\
\hline
VAIKAE & $\mathbf{0.0241}$ & $\mathbf{0.0473}$ \\
\hline
VIKAE & $0.0304$ & $0.0567$ \\
\hline
VKAE & $0.0276 $ & $0.0540$ \\
\hline
KAE ensemble & $0.0258$ & $0.0530$ \\
\hline
\end{tabular}
\end{table}

These models all have similar parameter counts except for the KAE ensemble, as each of its $16$ members has about as many parameters as non-ensembled models. The training dataset contains slices of the full time series, spanning at most $100$ time steps. 
Further implementation details can be found in appendix~\ref{sec:implementation_satellite}.
Our metric is 
the CRPS, computed on time steps $T_{train}$ to $T_{train} + T_{test}$ (for Fontainebleau) or $1$ to $T_{train} + T_{test}$ (for Orléans) with regard to the available observations in this range, using a single observation $\mathbf{y}_0$ at time 0 as input.
As summarized by table~\ref{tab:Sentinel_forecasting}, VAIKAE performs best on both spatial areas. Consistently with the results of~\cite{frion2025augmented} in a deterministic setting, VIKAE performs worst. This can be interpreted as a consequence of the restricted latent dimension $d=n$ of an invertible model, which prevents it from finding an accurate Koopman invariant subspace for this system. VKAE, which does not face this restriction, performs better than VIKAE, yet VAIKAE is the only ensemble-free model that outperforms the KAE ensemble.

\begin{figure}
\begin{center}
    \includegraphics[width=\linewidth]{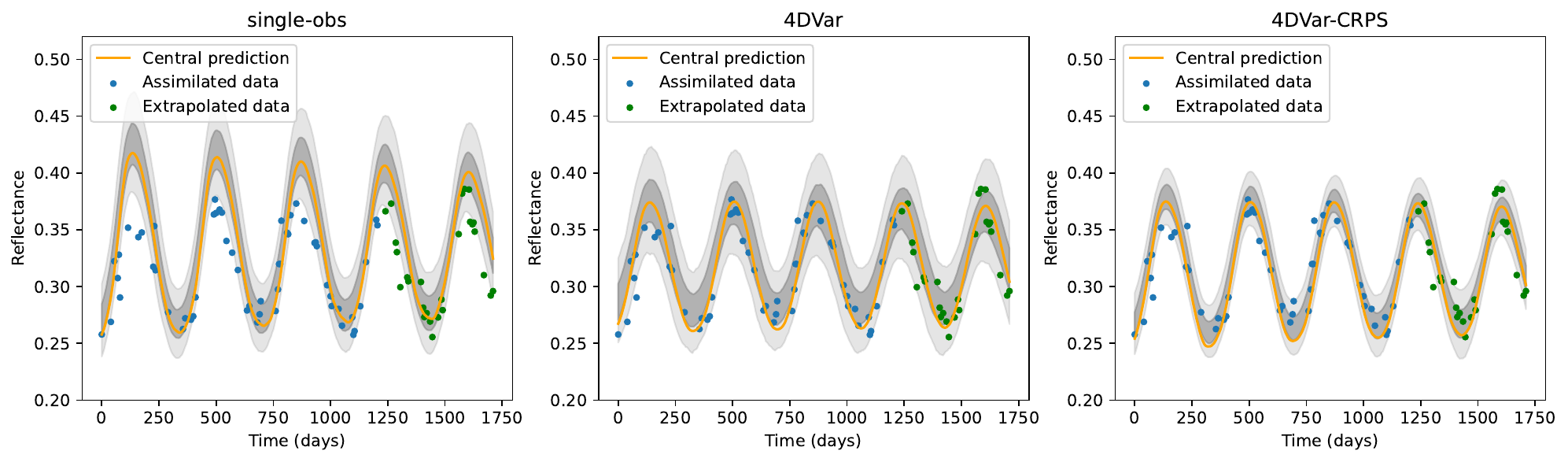}
\end{center}
\caption{Probabilistic performance of three methods, the first of which leverages only the initial observation while the two others assimilate all the blue-colored datapoints. We show only the B7 band, from the infrared domain, which is the most energetic in the dataset, yet we remind that our methods jointly manipulate the 10 spectral bands of pixelwise reflectance vectors. The dark shadings represent 50\% confidence intervals while the light shadings represent 90\% confidence intervals.}
\label{fig:Fontainebleau_comparison}
\end{figure}

\subsection{Latent data assimilation with a trained model}
\label{sec:satellite_DA}

\begin{figure}
\begin{center}
    \includegraphics[width=\linewidth]{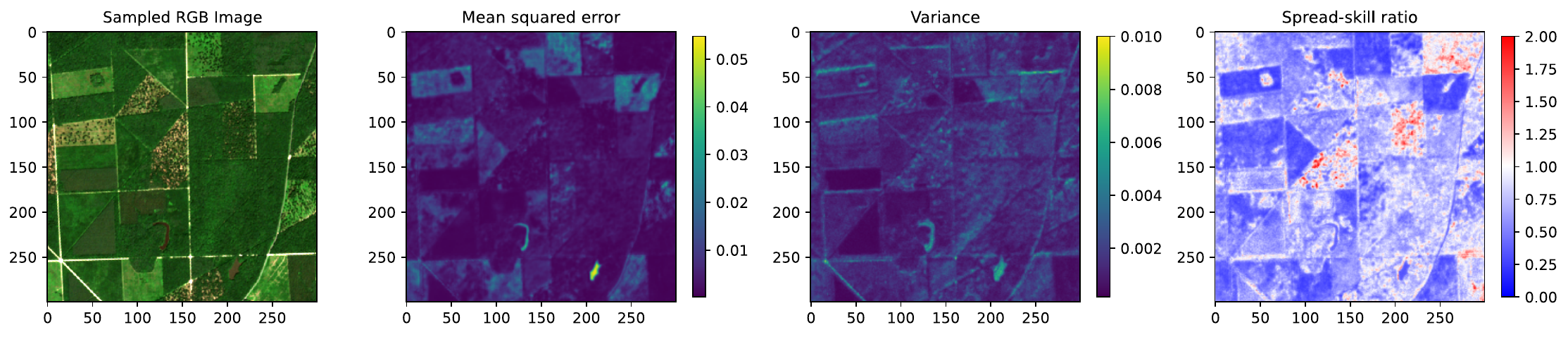}
\end{center}
\caption{Summary of 4DVar-CRPS extrapolation results on the Orléans area. From left to right: 1) RGB composition of a true image 2) Mean squared error on the B7 spectral band, averaged over the extrapolation time range 3) Corresponding variance 4) Corresponding spread-skill ratio. Since the ideal value is 1, white marks a good calibration while blue and red respectively indicate overconfidence and underconfidence.}
\label{fig:orleans_spsk}
\end{figure}

Having established the superiority of VAIKAE to other KAE architectures for probabilistic long-term forecasting from a single observed reflectance vector, we now study how such a trained model can be used to leverage multiple observations with data assimilation, using the methods presented in section~\ref{sec:methods_assimilation}. We consider 3 different procedures for predicting the state of the system from time $T_{train}$ to $T_{train} + T_{test}$:
\begin{itemize}
    \item Predicting without assimilation, using only $\mathbf{y}_0$ as an input, corresponding to the evaluation criterion in section~\ref{sec:Sentinel_training}. We refer to this method as \textbf{single-obs}.
    \item Assimilating on all available observations $\mathbf{y}_t$ such that $t$ is between 0 and $T_{train}-1$, using equations~\ref{eq:LDA_method1_eq1} and~\ref{eq:LDA_method1_eq2}. We refer to this method as \textbf{4DVar}.
    \item Assimilating the same set of observations with~\eqref{eq:LDA_method2}, thus jointly optimizing on the initial latent mean and variance. We refer to this method as \textbf{4DVar-CRPS}.
\end{itemize}

\begin{table}[tbp]
\centering
\caption[size=9pt]{Forecasting CRPS obtained when predicting from a single observation or assimilating on multiple observations with two different latent data assimilation methods.}
\label{tab:Sentinel_DA}
\begin{tabular}{|c|c|c|}
\hline
 & CRPS on the Fontainebleau area  & CRPS on the Orléans area \\
\hline
single-obs & $0.0241$ & $0.0457$  \\
\hline
4DVar & ${0.0153}$ & ${0.0273}$ \\
\hline
4DVar-CRPS & $\mathbf{0.0140}$ & $\mathbf{0.0244}$ \\
\hline
\end{tabular}
\end{table}

\begin{figure}
\begin{center}
    \includegraphics[width=\linewidth]{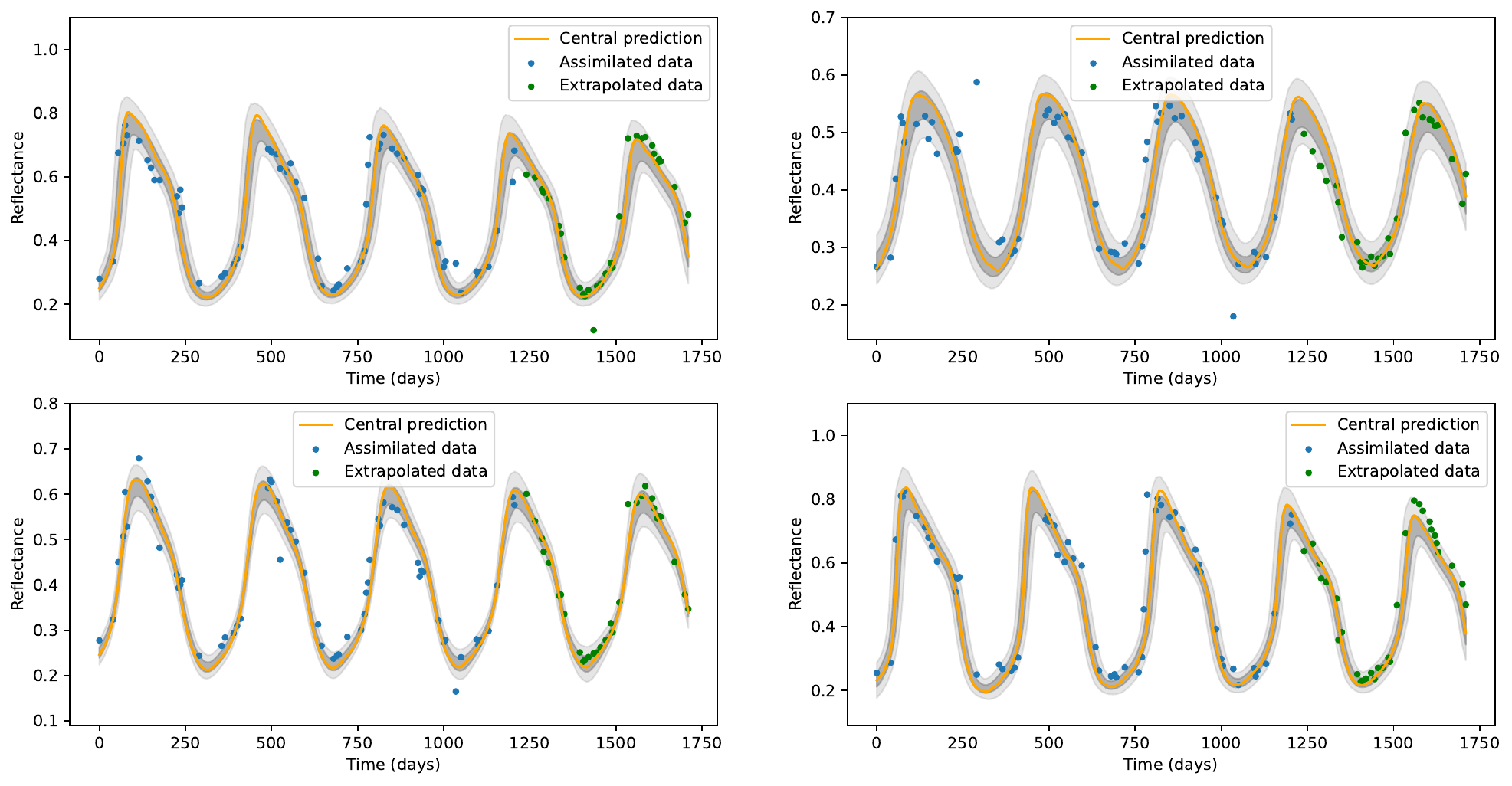}
\end{center}
\caption{4DVar-CRPS results on four randomly selected pixels from the test Orléans area. We consider the B7 spectral band, which is the most energetic in the dataset, although our method jointly processes the 10 spectral bands of the pixels. For each subfigure, the dark and light shading respectively correspond to the 50\% and 90\% confidence intervals of the predicted probability distributions.}
\label{fig:confidence_intervals_Orléans}
\end{figure}

We compute the CRPS obtained when estimating the reflectance vectors on time steps $T_{train}$ to $T_{train}+T_{test}$, which are located beyond the assimilation window. The results of all 3 methods, summarized in Table~\ref{tab:Sentinel_DA}, show that the CRPS in both areas is significantly reduced when multiple observations are included instead of a single one, marking a large improvement from single-obs to 4DVar. When using 4DVar-CRPS to fit both the mean and variance of the initial latent Gaussian distribution, a smaller additional gain is obtained. Interestingly, the performance gap between the training Fontainebleau area and the test Orléans area gets reduced, both in absolute and relative value, with stronger assimilation techniques. This shows that data assimilation can alleviate the distribution shift of the data, even without re-training or fine-tuning a model. 

\begin{wrapfigure}{h}{0.65\textwidth}
\begin{center}
    \includegraphics[width=0.65\textwidth]{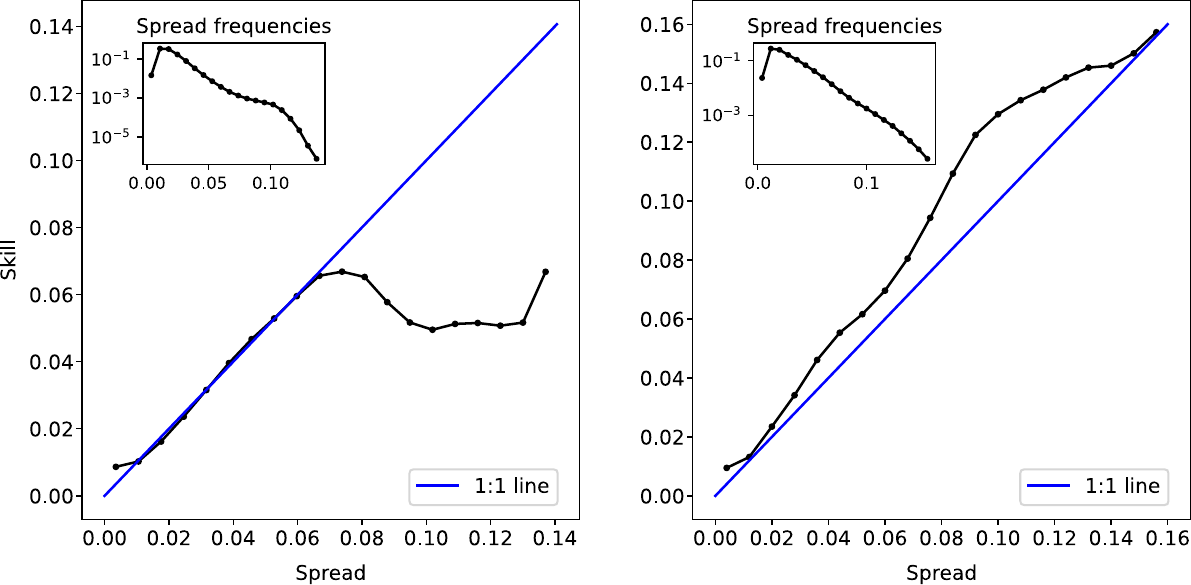}
\end{center}
\caption{Spread-skill plots for time series extrapolation with 4DVar-CRPS on the Fontainebleau (train) and Orléans (test) areas.}
\vspace{-8pt}
\label{fig:spread_skill_Fontainebleau_Orléans}
\end{wrapfigure}

To bring more visual insight into the behavior of the three methods, we show their respective predictions on a randomly selected pixel of the Fontainebleau area in figure~\ref{fig:Fontainebleau_comparison}. From this example, one can see that single-obs, despite having some skill, does not accurately fit the local maxima of the reflectance dynamics, either in the assimilation or extrapolation domain. In contrast, 4DVar's central prediction better captures the true dynamics, including in the extrapolation range, yet it is clearly underconfident. 4DVar-CRPS corrects this flaw by tightening the uncertainties around a similarly-behaving central prediction. On figure~\ref{fig:orleans_spsk}, we display temporally aggregated statistics of our 4DVar-CRPS predictions over the test Orléans area, again focusing on the most energetic spectral band of the dataset. One can see that the spread-skill ratio (SSR) remains close to its ideal value of~1 in a large part of this spatial domain, but that some areas have largely overconfident predictions with a SSR below 1 while a smaller portion of the domain has underconfident predictions with a SSR above~1. We additionally display some 4DVar-CRPS confidence intervals on the test Orléans area in figure~\ref{fig:confidence_intervals_Orléans}, qualitatively showing well calibrated predictions. These results are further discussed and illustrated with corresponding sampled state trajectories in appendix~\ref{sec:additional_Orléans}. 

Finally, 
figure~\ref{fig:spread_skill_Fontainebleau_Orléans} shows spread-skill plots~\citep{haynes2023creating} of the 4DVar-CRPS predictions on both spatial areas in the extrapolation time domain. These plots are obtained by binning all predictions (mixing time steps, spatial locations and spectral bands) in a histogram according to their spread (i.e. standard deviation) estimated over many samples, and plotting the average skill (i.e. root mean squared error) for each bin. The inset spread frequencies plot indicates the relative sizes of the spread bins. A perfect spread-skill plot would match the 1:1 line, as the spread should match the skill on average. On the figure, one can see that the Fontainebleau predictions are nearly perfectly calibrated for low spread values but underconfident for large spread values, which however correspond to much fewer datapoints, and thus have a limited impact on the global SSR, which has a nearly perfect value of 1.04. The predictions on Orléans are overall slightly overconfident, leading to a global SSR of 0.83.

\section{Conclusion}
\label{sec:conclusion}

In this paper, we have introduced VAIKAE, a Koopman autoencoder model that enables probabilistic time series forecasting with a linearly evolving Gaussian distribution as its latent dynamics. We showed that this model enables analytical likelihood computation for the state at subsequent times given an observed initial state, thus inspiring a likelihood-based training criterion. We then discussed how to leverage a pre-trained VAIKAE model for uncertainty-aware latent data assimilation, enabling long-term time series forecasting using multiple irregularly-sampled state observations. We finally demonstrated the strong performance of our methods, with well-calibrated predictions on two long-term time series forecasting benchmarks, one of which contains irregularly-sampled time series. 

Interesting directions for future work might include training our model with a CRPS-based loss criterion, as is becoming increasingly popular for global atmosphere forecasting models. This approach has an important computational cost compared to deterministic forecasting, but scales better than our likelihood-based approach for large state dimensions. Besides, our new data assimilation method for optimizing jointly on the mean and variance of the initial latent state using a CRPS-based variational cost should be explored in more general contexts than for VAIKAE only.

\bibliography{main}
\bibliographystyle{tmlr}

\newpage
\appendix

\section{Representation of the AIKAE architecture}
\label{sec:AIKAE}

\begin{figure}
\begin{center}
    \includegraphics[width=\linewidth]{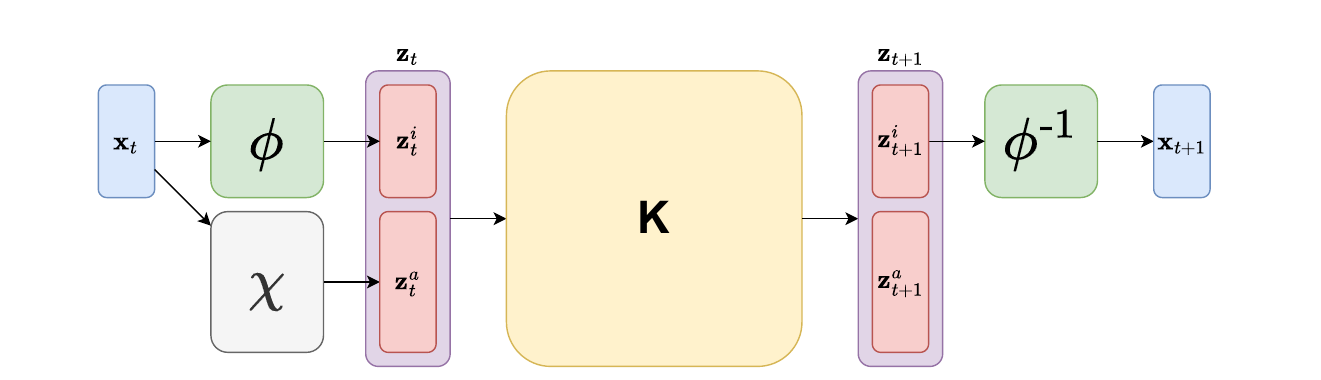}
\end{center}
\caption{Visual representation of the AIKAE architecture, copied with permission from~\cite{frion2025augmented}.}
\label{fig:AIKAE}
\end{figure}

The VAIKAE architecture, which is presented in section~\ref{sec:methods_VAIKAE}, is based upon the AIKAE architecture introduced by~\cite{frion2025augmented}. We show a visual representation of the AIKAE in figure~\ref{fig:AIKAE}. The main difference between the new VAIKAE architecture shown in figure~\ref{fig:VAIKAE} and the AIKAE is that the size of the output of $\chi$ is increased so that it produces a latent diagonal variance vector $\boldsymbol{\sigma}_t$ for the whole latent distribution in addition to the mean $\boldsymbol{\mu}_t^a$ of the augmentation part of the latent space.

\section{Description of the datasets}
\label{sec:datasets}

We consider 6 standard datasets for long-term time series forecasting, with varying sizes and properties. The \textit{ETTm1} and \textit{ETTm2} datasets~\citep{zhou2021informer} correspond to measurements of 7 factors from electricity transformers, such as load and oil temperature, recorded every 15 minutes from July 2016 to July 2018. \textit{Weather}~\citep{wu2021autoformer} contains 21 weather indicators recorded every 10 minutes by the Max Planck Biogeochemistry Institute in 2020. \textit{Solar} (\cite{lai2018modeling}, also called Solar-Energy) contains measurements from the solar power production of 137 photovoltaic plants, recorded every 10 minutes in 2006. \textit{ECL}~(\cite{wu2021autoformer}, also called Electricity) reports the hourly electricity consumption of 321 clients from 2012 to 2014. Finally, \textit{Traffic}~\citep{wu2021autoformer} records the hourly occupancy rates of 862 roads in the San Francisco Bay Area, from January 2015 to December 2016.

For all of these datasets, we adopt the same approach as~\cite{li2025diffusion} and many other previous works by using the earlier time steps of the time series for training, an intermediary time range for validation and the latest part for testing. The length ratios of the train-validation-test splits are 6:2:2 for the ETTm1 and ETTm2 datasets, and 7:1:2 for all other datasets.

\section{Description of the evaluation metrics}
\label{sec:metrics}

\subsection{Deterministic metrics}

\textbf{Mean squared error (MSE)} is commonly used as a training criterion and as an evaluation metric for neural networks on regression problems. It is defined as the averaged squared error between a pointwise prediction (or, alternatively, the mean of a predicted distribution) and the corresponding groundtruth values. For a time series $(\mathbf{x}_t^i)_{1 \leq t \leq T, 1 \leq i \leq n}$ with an $n$-dimensional state $\mathbf{x}_t \in \mathbb{R}^n$ and a corresponding prediction $(\hat{\mathbf{x}}_t^i)_{1 \leq t \leq T, 1 \leq i \leq n}$, the MSE is computed as:
\begin{equation}
    \textrm{MSE} = \frac{1}{nT} \sum_{i=1}^n \sum_{t=1}^T (\mathbf{x}_t^i - \hat{\mathbf{x}}_t^i)^2.
\end{equation}

\textbf{Mean average error (MAE)} corresponds to the average absolute difference between the predicted and true state, over different variables and time steps of a time series. With similar notation as above, it can be expressed as:
\begin{equation}
    \textrm{MAE} = \frac{1}{nT} \sum_{i=1}^n \sum_{t=1}^T |\mathbf{x}_t^i - \hat{\mathbf{x}}_t^i|.
\end{equation}

\subsection{Proper scoring rules for probabilistic forecasts}
\label{sec:proper_scoring_rules}

Proper scoring rules~\citep{gneiting2007strictly} are central to the training and evaluation of data-driven stochastic models. In short, proper scoring rules are used to compare a predicted probability distribution to a single true datapoint that is supposedly sampled from a groundtruth probability distribution. In the limit of an infinite number of datapoints, (strictly) proper scoring rules are minimized (only) when the predicted probability distribution is the same as the groundtruth distribution. A popular proper scoring rule for both training and evaluating probabilistic neural networks is the negative log-likelihood, also called the logarithmic score (see e.g.~\cite{lakshminarayanan2017simple,kendall2017uncertainties}). However, some pitfalls of negative log-likehood as a training criterion have been identified by e.g.~\cite{skafte2019reliable,seitzer2022on}. This partly explains the recent gain in popularity of the CRPS, which we present next.

\textbf{Continuous ranked probability score (CRPS)} is a strictly proper scoring rule, defined for a univariate probability distribution, described by its cumulative distribution function $F$. When the true observed value is $x_{true} \in \mathbb{R}$, the CRPS can be written as:
\begin{equation}
    \textrm{CRPS} = \int_\mathbb{R} (F(x) - \mathds{1}_{x \geq x_{true}})^2dx,
\end{equation}
where $\mathds{1}_{x \geq x_{true}}$ denotes the indicator function, with value 1 when $x \geq x_{true}$ and 0 otherwise. Importantly, for multivariate quantities, one usually considers the average of the CRPS over all variables of the state, although this means evaluating the marginals of this multivariate distribution rather than the joint distribution~\citep{alet2025skillful}.

In practice, as detailed by e.g.~\cite{gneiting2007strictly}, when the predicted probability distribution is defined (or approximated) by a finite ensemble of $M$ equiprobable values $x_1, ..., x_M$, the CRPS can be written as
\begin{equation}
\label{eq:crps}
    \text{CRPS} = \frac{1}{M}\sum_{i=1}^M |x_{true}-x_i| - \frac{1}{2}\frac{1}{M^2}\sum_{j=1}^M\sum_{k=1}^M |x_j - x_k|.
\end{equation}
This expression is easy to compute and differentiate in practice, and enables estimation of the CRPS of a complex probability distribution by drawing $M$ samples. It has thus been used to train multiple probabilistic forecasting models, notably for recent data-driven models of the atmosphere~\citep{price2025probabilistic,alet2025skillful,lang2026aifs,agarwal2026skillful}. In this work, we do not use this strategy for training our models, although this would be a possibility. We however use the expression from~\eqref{eq:crps} in order to solve a gradient descent on the latent variational data assimilation cost of~\eqref{eq:LDA_method2}.

\section{Implementation details}

\subsection{Informer benchmark}
\label{sec:implementation_Informer}

For all datasets in this benchmark, the invertible encoder $\phi$ is implemented using a NICE~\citep{dinh2014nice} normalizing flow model with $l$ coupling layers, each using the additive coupling law with a simple multi-layer perceptron (MLP) comprising a single layer with width 256 and a leaky rectified linear unit nonlinearity~\citep{maas2013rectifier}. The second encoder $\chi$ is simply an MLP network with 3 layers of respective widths [256, 128, 160]. Note that, as outlined in section~\ref{sec:methods_VAIKAE}, the output layer is of size $160 = 128 + 32 = p + d = n + 2d$, as the input to the model is of size $n=96$ and we have fixed the size of the augmentation encoding to $d=32$. 

Like several strong variable-independent methods~\citep{li2023revisiting,frion2025augmented}, we use a reversible instance normalization (RevIN,~\cite{kim2021reversible}) of the input. We use the training loss of~\eqref{eq:complete_loss} with relative weights $\alpha=\beta=0, \gamma=10^{-2}$. The models are trained with the Adam optimizer, using a learning rate $r$ and default momentum parameters $\beta_1 = 0.9,  \beta_2 = 0.999$. The batch size is set to $128$. A hyperparameter search was performed for each dataset on the following values: $l \in \{4, 6\}, r \in \{10^{-3}, 2 \cdot 10^{-3}\}$.

In our experiments on this benchmark, we observed a trade-off between accurate prediction of the conditional mean and uncertainty calibration, controlled by the weight used for $\gamma$ in the loss function. $\gamma = 0$ reduces to the AIKAE~\citep{frion2025augmented} in practice, which is a strong model for deterministic predictions but does not provide uncertainties. As $\gamma$ increases, the uncertainties increase (i.e. become better calibrated) yet the deterministic MSE and MAE metrics tend to simultaneously worsen. We solve this dilemma by choosing the relatively low $\gamma = 10^{-2}$, which leads to a highly overconfident model, and then performing a recalibration of the uncertainties as a post-processing method. Concretely, for each variable and prediction time step, we compute the average spread-skill ratio over the validation dataset. Afterwards, when evaluating the model's predictions, we inflate the variable-and-time-step-wise standard deviation of the predictions with the inverse of these observed spread-skill ratios.

\subsection{Satellite image time series benchmark}
\label{sec:implementation_satellite}

In contrast with the other set of experiments, our VAIKAE model now uses the Real-NVP normalizing flow architecture~\citep{dinh2017density}, which is significantly more expressive than the previously used NICE architecture~\citep{dinh2014nice} but more prone to instabilities. We stack 6 Real-NVP coupling layers, which each internally uses a simple MLP network with a single hidden layer of width 256. The augmentation encoder $\chi$ is a MLP network with 2 hidden layers of respective sizes 512 and 256. The augmentation part $\mathbf{z}_t^a$ of the latent embedding is of size $p=16$. Since the state is a vector of $n=10$ reflectance values, the full latent size of the model is $d = n + p = 26$.

\begin{figure}
\begin{center}
    \includegraphics[width=15cm]{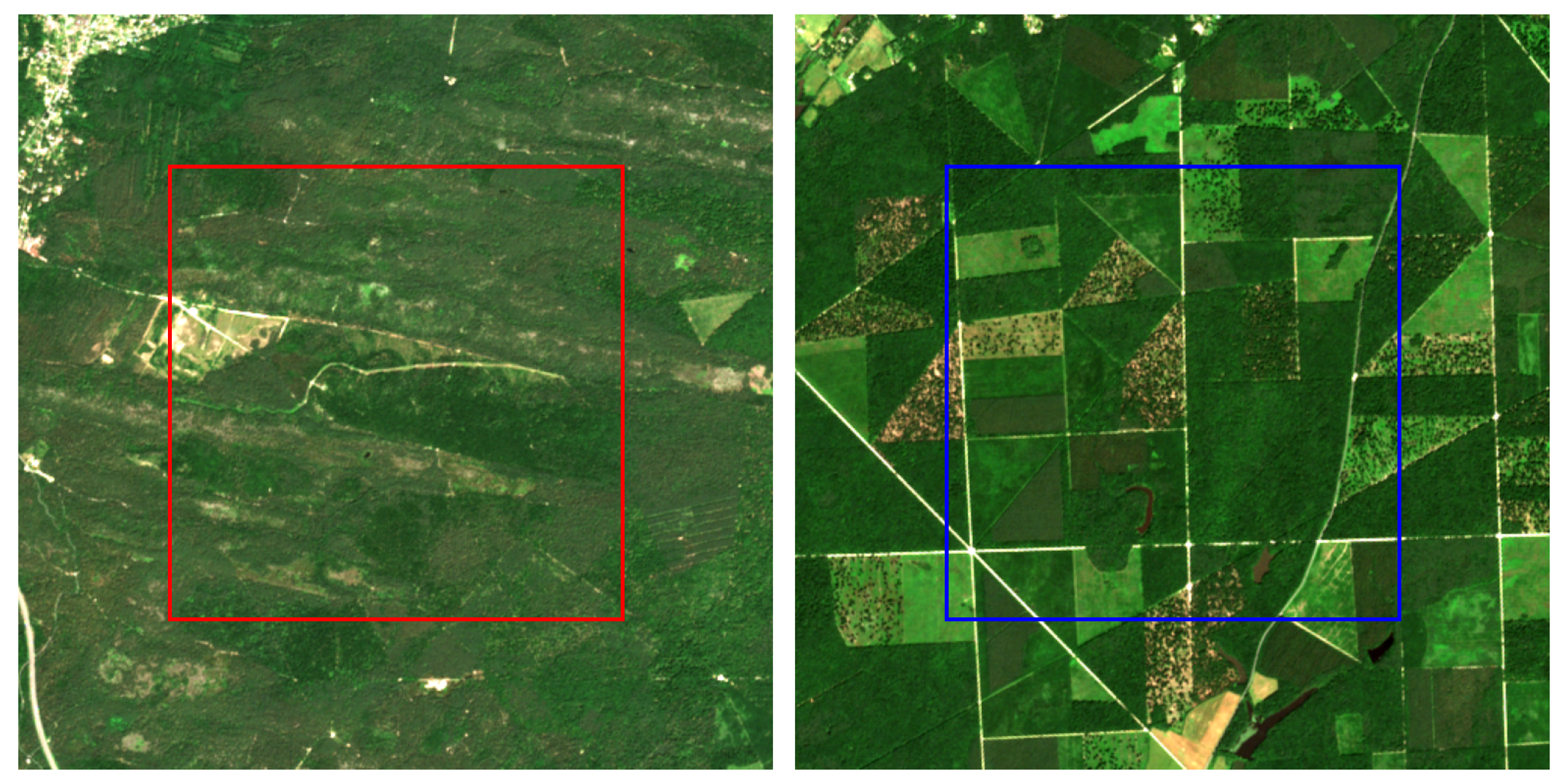}
\end{center}
\caption{Left: an image from the forest of Fontainebleau. Right: an image from the forest of Orléans. The date for both images is 20/06/2018. Those are RGB compositions with saturated colors. The red square is the $300 \times 300$ pixel training area and the blue square is the $300 \times 300$ pixel test area.}
\label{fig:Sentinel_sample}
\end{figure}

On figure~\ref{fig:Sentinel_sample}, we show RGB compositions of images from the two considered spatial areas, along with the subdomains used for training and testing our models. As mentioned in the main text, the whole training domain is of size $242 \times 300 \times 300 \times 10$, with $242$ time steps, $300 \times 300$ pixels and $n=10$ spectral bands, yet the observations are sparse since only about $1$ in $5$ time steps are actually observed, although all pixels and spectral bands are always observed at the same time. We create slices of observations starting from each observed pixel before time $T=142$ and then including all available observations for the subsequent $100$ time steps. These slices of observations are randomly separated into batches of size $2048$ and used for training with the loss of~\eqref{eq:complete_loss}. For VAIKAE and VIKAE, we use $\alpha=10^{-3}, \beta=10^{-2}, \gamma=10^5$. We use the Adam optimizer with a learning rate of $10^{-4}$. For VKAE, since the reconstruction is not guaranteed to be exact by design, we additionally use a reconstruction loss term as in~\cite{frion2024neural}, and the likelihood loss term is adapted so that the intractable determinant of the Jacobian of the latent embedding is simply substituted by~$1$, which amounts to ignoring the deformation of the latent space. Our validation criterion is the mean squared error of predictions from time~$0$ to~$242$ on a subset of the spatial training domain. Despite the relatively low value of the weight $\lambda$ for the likelihood loss terms, the uncertainties are relatively well calibrated at the end of the training, and a post-processing recalibration is not required. 

\section{Additional satellite image time series results}
\label{sec:additional_Orléans}

\begin{figure}
\begin{center}
    \includegraphics[width=\linewidth]{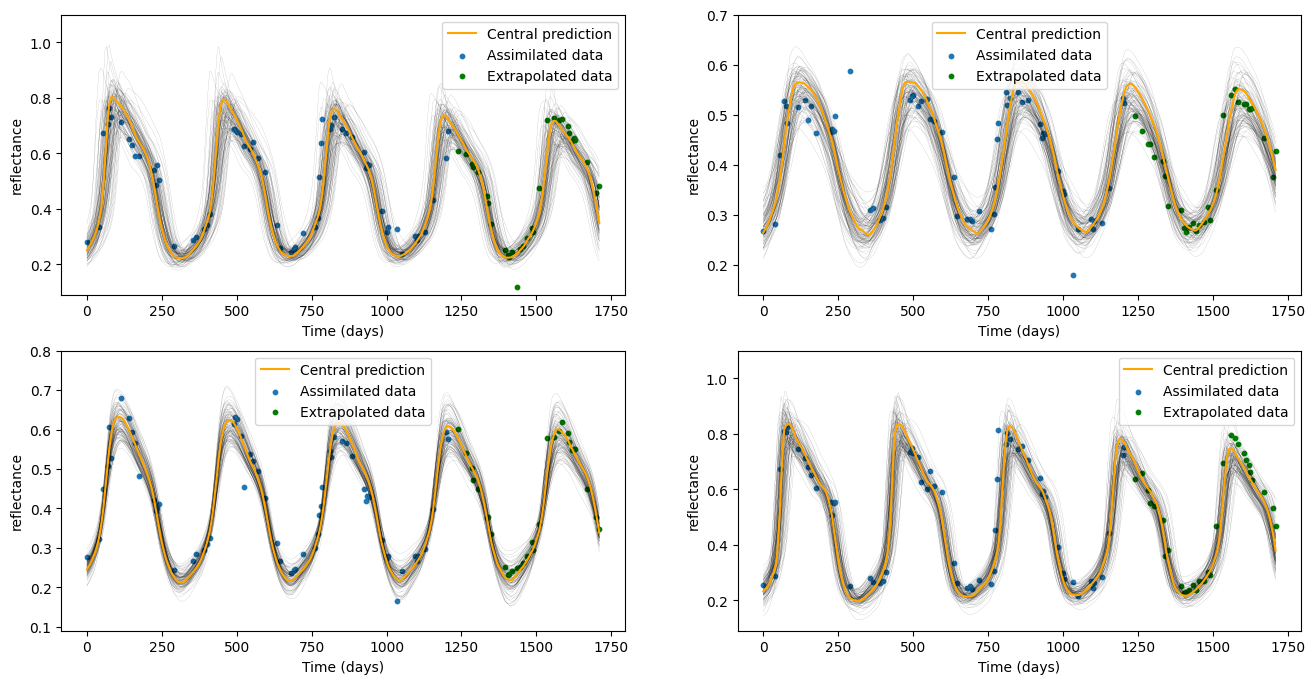}
\end{center}
\caption{Each subfigure corresponds to a randomly selected pixel from the Orléans area, for which we assimilate the blue-colored observations and sample 100 state trajectories from the inferred distribution.}
\label{fig:Sentinel_trajectories}
\end{figure}

Here, we plot some results obtained with the "4DVar-CRPS" assimilation method, described in~\eqref{eq:LDA_method2}, which jointly optimizes on the mean and variance of the initial latent Gaussian embedding of the state. We randomly sample 4 pixels from the Orléans area, from which no data was observed while training the VAIKAE model. On figure~\ref{fig:Sentinel_trajectories}, we show for each of these pixels 100 sampled trajectories from the assimilated initial distribution, including the "central prediction" which corresponds to the mean of the initial latent Gaussian distribution but not necessarily to the mean or median of the trajectories in the state space, due to the deformation between the latent and state spaces. These predictions are performed over the same data as for the confidence intervals of figure~\ref{fig:confidence_intervals_Orléans}, and in fact illustrate how such confidence intervals are obtained: by sampling a large number of trajectories and computing associated quantiles for each variable and time step. From figure~\ref{fig:confidence_intervals_Orléans}, one can see that most of the observations are captured by the 90\% confidence interval, although some clear outliers, either in the assimilation or extrapolation range, remain outside of it, which does correspond to the expected behavior of the probabilistic forecasting model. This visually relevant identification of seemingly anomalous values in the time series hints at how a trained VAIKAE model could be used alongside with 4DVar-CRPS for anomaly detection.

\end{document}